\documentclass{article}

\usepackage[main, final]{neurips_2026}

\usepackage[utf8]{inputenc} 
\usepackage[T1]{fontenc}    
\usepackage{hyperref}       
\usepackage{url}            
\usepackage{booktabs}       
\usepackage{amsmath}        
\usepackage{amsfonts}       
\usepackage{amssymb}        
\usepackage{nicefrac}       
\usepackage{microtype}      
\usepackage[table]{xcolor}         
\usepackage{array}
\usepackage{subcaption}

\newcommand{\muold}{\mu_{\mathrm{old}}}

\usepackage{color}
\usepackage{amsmath}
\usepackage{amsthm}
\usepackage{makecell} 
\usepackage{graphicx}

\title{Missing Old Logits in Asynchronous Agentic RL: Semantic Mismatch and Repair Methods for Off-Policy Correction}

\author{%
  Zhong Guan$^{1}$\thanks{Equal contributions.},
  Yongjian Guo$^{2,4}$\footnotemark[1],
  Haoran Sun$^{3,4}$\footnotemark[1],
  Wen Huang$^{2,4}$\footnotemark[1], \\
  Shuai Di$^{4}$,
  Likang Wu$^{1}$, \\
  Xiong Jun Wu$^{4}$\thanks{Corresponding author. (\texttt{xiongjunwu.1@jd.com})},
  Hongke Zhao$^{1}$\thanks{Corresponding author. (\texttt{hongke@tju.edu.cn})} \\
  \\
  $^1$Tianjin University, $^2$Tsinghua University, $^3$Peking University, $^4$JDT AI Infra
}

\begin{document}

\maketitle

\begin{abstract}

Asynchronous reinforcement learning improves rollout throughput for large language model agents by decoupling sample generation from policy optimization, but it also introduces a critical failure mode for PPO-style off-policy correction. In heterogeneous training systems, the total importance ratio should ideally be decomposed into two semantically distinct factors: a \emph{training--inference discrepancy term} that aligns inference-side and training-side distributions at the same behavior-policy version, and a \emph{policy-staleness term} that constrains the update from the historical policy to the current policy. We show that practical asynchronous pipelines with delayed updates and partial rollouts often lose the required historical training-side logits, or old logits. This missing-old-logit problem entangles discrepancy repair with staleness correction, breaks the intended semantics of decoupled correction, and makes clipping and masking thresholds interact undesirably. To address this issue, we study both exact and approximate correction routes. We propose three exact old-logit acquisition strategies: snapshot-based version tracking, a dedicated old-logit model, and synchronization via partial rollout interruption, and compare their system trade-offs. From the perspective of approximate correction, we focus on preserving the benefits of decoupled cor rection through a more appropriate approximate policy when exact old logits cannot be recovered at low cost, without incurring extra system overhead. Following this analysis, we adopt a revised PPO-EWMA method, which achieves significant gains in both training speed and optimization performance.
\end{abstract}

\section{Introduction}
\label{sec:introduction}

Large-scale reinforcement learning for large language models (LLMs) increasingly relies on distributed rollout and training pipelines. Proximal Policy Optimization (PPO)~\citep{schulman2017proximal} and its variants~\citep{yu2025dapo,qi2026rethinking,ahmadian2024back,ahmadian2024basicsrevisitingreinforcestyle}, including GRPO~\citep{shao2024deepseekmath}, remain widely used because they provide a simple and stable mechanism for policy improvement: trajectories are generated by a behavior policy, and the current policy is optimized with an importance ratio and a clipped surrogate objective. In ideal on-policy or near-synchronous settings, this ratio has a clear interpretation. It compares the current policy against the policy that generated the sampled tokens, while clipping trades off update magnitude and optimization stability.

This interpretation becomes fragile in modern Agentic RL systems. To maximize throughput, rollout and training are often physically separated. Rollouts are produced by optimized inference engines such as vLLM~\citep{kwon2023efficient} or SGLang~\citep{zheng2024sglang}, whereas gradient updates are performed by training engines such as Megatron-LM or FSDP. Even when the inference and training sides nominally use the same model version, numerical kernels, precision scaling, quantization, tensor parallelism, and routing implementations can lead to different token probabilities. We call this effect \emph{training--inference discrepancy}~\citep{yao2025offpolicy}. At the same time, asynchronous rollouts, large rollout queues, partial trajectories, and multiple actor updates make the behavior policy stale with respect to the current policy. We call this effect \emph{policy staleness}.

A natural choice for correction is to decompose the total ratio into two terms: a discrepancy-repair ratio that compares the training-side and inference-side distributions at the same old version, and a staleness-correction ratio that compares the current training policy with that old training-side policy~\citep{xiao2026mimo,team2026kimi,zeng2026glm,wang2026ernie,team2025every}. 
Let $\mu_{\mathrm{old}}$ denote the inference-side rollout policy, and let $\pi_{\mathrm{old}}$ denote the corresponding training-side forward policy. The desired decomposition is
\begin{equation}
    r(\theta)
    =
    \frac{\pi_{\theta}(y|x)}{\mu_{\mathrm{old}}(y|x)}
    =
    r_d r_s,
    \quad
    r_d
    =
    \frac{\pi_{\mathrm{old}}(y|x)}{\mu_{\mathrm{old}}(y|x)},
    \qquad
    r_s
    =
    \frac{\pi_{\theta}(y|x)}{\pi_{\mathrm{old}}(y|x)} .
    \label{eq:intro_decomposition}
\end{equation}
Here $r_d$ measures training--inference discrepancy, while $r_s$ measures policy staleness. This decomposition is attractive because the two terms have different meanings and should be controlled differently. Discrepancy repair should filter or down-weight numerically inconsistent tokens. Staleness correction should constrain policy updates with the sign-dependent PPO clipping rule.

However, asynchronous Agentic RL~\citep{dong2025agentic,wang2025let,zhanglandscape} introduces a practical obstacle: the old training-side policy values $\pi_{\mathrm{old}}(y|x)$ may no longer be available when the trajectory reaches the actor. 
This is especially common under partial rollout collection, where one trajectory can span multiple parameter versions, and the actor may already have advanced beyond the version that generated earlier tokens. Once these old logits are missing, the decomposition in Eq.~\eqref{eq:intro_decomposition} is no longer semantically valid. Existing decoupled objectives may then mix discrepancy repair and staleness correction into a proxy ratio, causing the clipping and masking mechanisms to interfere with each other. 
However, current training stacks, including Verl~\citep{sheng2025hybridflow}, ROLL~\citep{wang2025reinforcement}, and SLIME~\citep{slime_github}, still leave the old-logit mismatch unresolved.

\begin{figure}[t]
    \centering
    \includegraphics[width=1\linewidth]{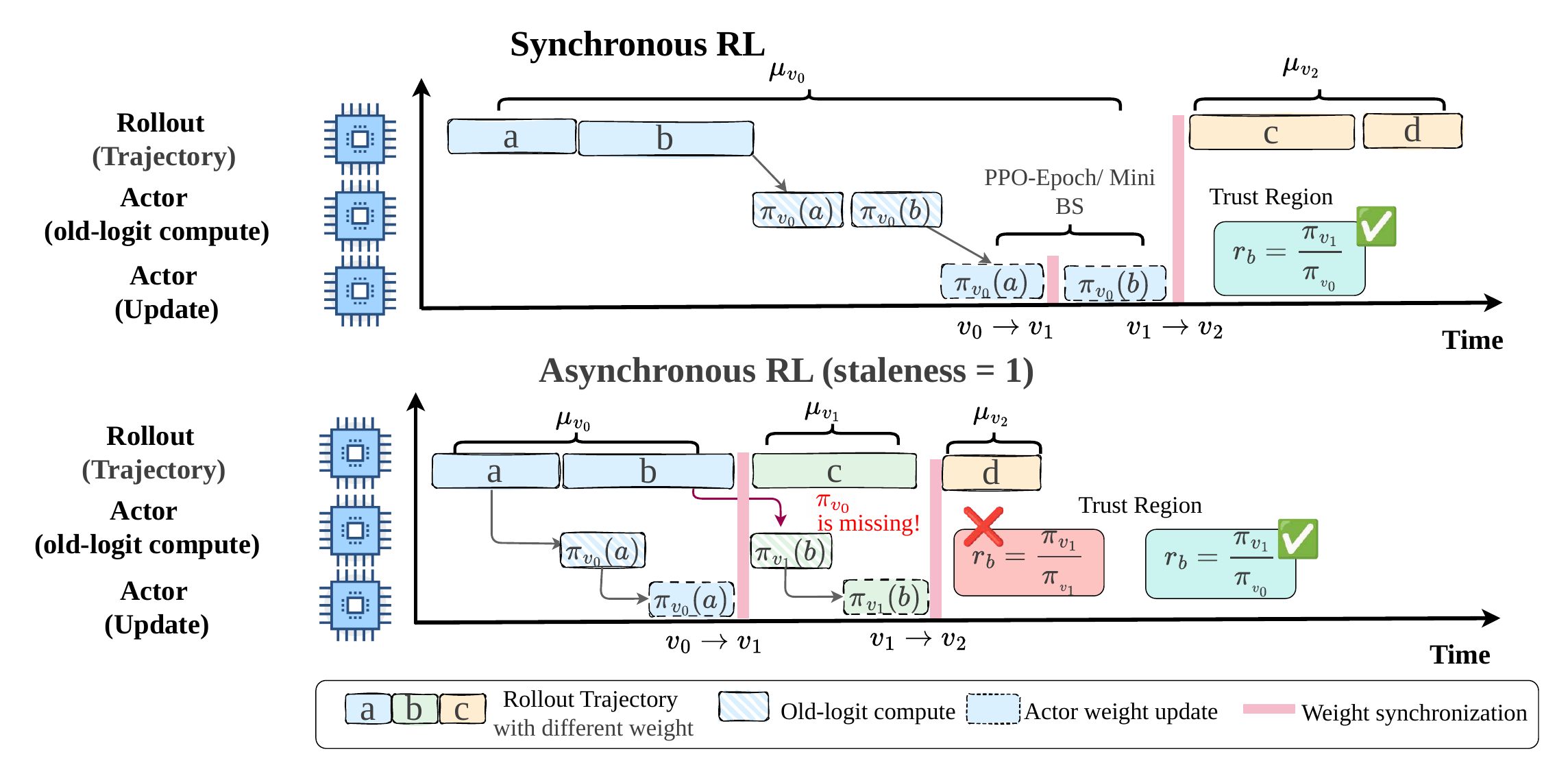}
    \caption{Synchronous versus asynchronous RL. In synchronous RL, the old logits used during training correspond to the same policy version that generated the rollout. In asynchronous RL, delayed updates and partial rollouts can make this version unavailable, which creates an old-logit mismatch and breaks the intended separation between training--inference discrepancy repair and policy-staleness correction.}
    \label{fig:intro}
\end{figure}

This paper studies the missing-old-logit problem in asynchronous LLM RL. We first give a unified view of existing objectives as imposing two distinct constraints: a discrepancy constraint and a staleness constraint. This view shows why using one ratio or one threshold for both effects is insufficient. We then analyze interpolation-based proxy policies and show that, under common constructions, they mainly re-parameterize effective clipping boundaries rather than recovering the missing reference policy. Finally, we examine two practical directions: exact acquisition of old logits through system support, and low-cost approximation through an exponentially-weighted moving average PPO (PPO-EWMA) reference policy~\citep{hilton2022batch}.

Our contributions are summarized as follows.
\begin{itemize}
    \item \textbf{We identify the missing-old-logit problem in asynchronous Agentic RL.} Missing training-side old logits break the intended separation between training--inference discrepancy repair and policy-staleness correction, creating a semantic failure mode in decoupled correction objectives.
    \item \textbf{We provide a unified analysis and practical correction strategies.} We formulate existing PPO-style objectives under a dual-constraint view, clarify the need to decouple discrepancy repair from staleness correction, and show that interpolation-based proxies mainly re-parameterize clipping boundaries. We further study three exact old-logit acquisition routes and a revised PPO-EWMA reference as a low-cost approximation.
    \item \textbf{We evaluate the performance--cost trade-off on dense and MoE LLMs.} Experiments on Agentic benchmarks compare exact recovery, proxy references, and PPO-EWMA across optimization behavior and system overhead.
\end{itemize}

\section{Related Work}
\label{sec:related_work}

\subsection{Off-Policy and Asynchronous Reinforcement Learning for LLMs}

PPO~\citep{schulman2017proximal} and GRPO~\citep{shao2024deepseekmath} are widely used in LLM reinforcement learning because their clipped objectives stabilize policy updates while remaining straightforward to implement at scale~\citep{yu2025dapo}. However, on-policy training can be inefficient for long-horizon Agentic tasks, where rollout generation is expensive and GPU utilization is often limited by synchronization~\citep{guan2026rl}. This has motivated off-policy and asynchronous RL pipelines that reuse stale trajectories and decouple rollout generation from policy optimization.

Several recent methods~\citep{chen2025minimax,su2025klear} improve off-policy robustness by modifying the importance-sampling weights. CISPO~\citep{chen2025minimax} clips or regularizes importance weights for long-sequence training. GPPO~\citep{su2025klear} separates gradient propagation from clipping constraints to preserve useful exploratory gradients. M2PO~\citep{zheng2025prosperity} controls the second moment of importance weights to reduce variance under stale data. VESPO~\citep{shen2026vespo} and VCPO~\citep{huang2026stable} use effective sample size as a stability signal, while MiniRL~\citep{zheng2025stabilizing} and TOPR~\citep{roux2025tapered} modify trajectory-level importance weighting through tapered or asymmetric weighting. System-oriented work such as AReaL~\citep{fu2025areal}, HybridFlow~\citep{sheng2025hybridflow}, and related asynchronous frameworks study how to overlap rollout and training clusters at large scale.

These works demonstrate that off-policy and asynchronous training can substantially improve throughput. Our work focuses on a complementary issue: in heterogeneous asynchronous LLM RL, the policy version needed for a clean correction may be missing. This makes the meaning of the importance ratio ambiguous even before variance control or clipping design is considered.

\subsection{Training-Inference Mismatch and Reference Policy Correction}

Training--inference mismatch arises when the inference engine that produces rollouts and the training engine that computes gradients implement slightly different numerical computations. The mismatch is especially visible in MoE models, where routing decisions can amplify small numerical differences~\citep{zheng2025group}. Existing approaches mitigate this instability through masking, clipping, or routing replay. Masked Importance Sampling (MIS)~\citep{liuli2025rlcollapse} masks tokens with severe training--inference divergence. IcePop~\citep{zhao2025small} combines bilateral clipping and token masking to reduce the effect of unstable low-probability tokens. Routing-replay methods such as R2~\citep{zheng2025stabilizing} and R3~\citep{ma2025stabilizing} align expert routing between rollout and training, thereby reducing MoE-specific discrepancy.

A separate line of work builds reference or proximal policies to stabilize stale updates. Decoupled PPO~\citep{zheng2025stabilizing} separates importance correction from proximal constraints, while A-3PO~\citep{li20253po} approximates the proximal policy via log-space interpolation to reduce overhead. PPO-EWMA-style references maintain a smoothed policy anchor. These methods motivate our decoupled view. Our key distinction is that we examine whether the reference policy is semantically correct in asynchronous systems. When exact old training-side logits are absent, a proxy reference can help, but remains an approximation rather than true recovery of Eq.~\eqref{eq:intro_decomposition}.

\section{Preliminaries}
\label{sec:preliminaries}

\subsection{PPO-Style Off-Policy Correction}
\label{sec:ppo_style_off_policy_correction}

We consider RL fine-tuning of an LLM on prompts $x\sim P$. Given a prompt $x$, a response $y=(y_1,\ldots,y_T)$ is sampled from an old policy $\pi_{\mathrm{old}}$. In this standard PPO notation, $\pi_{\mathrm{old}}$ denotes the behavior policy, and we do not yet distinguish the inference-side rollout distribution from the training-side forward distribution. A reward model or environment returns a scalar reward $R(x,y)$, and an advantage estimate $A_t$ is computed for each token or sequence.

For a token-level ratio $r_t(\theta)=\pi_\theta(y_t|x,y_{<t})/\pi_{\mathrm{old}}(y_t|x,y_{<t})$, the PPO clipped surrogate is
\begin{equation}
    \ell_{\mathrm{PPO}}(\theta)
    =
    \mathbb{E}_{t}
    \left[
    \min\left(
        r_t(\theta) A_t,
        \mathrm{clip}(r_t(\theta),1-\epsilon,1+\epsilon)A_t
    \right)
    \right].
    \label{eq:ppo_surrogate}
\end{equation}
Equivalently, PPO clipping induces an advantage-sign-dependent active region. For $A_t>0$, ratios above $1+\epsilon$ are clipped; for $A_t<0$, ratios below $1-\epsilon$ are clipped.
Therefore, at the per-token level, we rewrite the gradient contribution of Eq.~\eqref{eq:ppo_surrogate} into a masked importance sampling (MIS) form: $\mathbb E_t[\mathrm{MIS} \cdot r_t(\theta) A_t \nabla \log \pi_\theta(y_t|x,y_{<t})]$, where the PPO-side active mask is defined as
\begin{equation}\label{eq:ppo_active_mask}
    \mathrm{MIS} 
    =
    \mathbb{I}\{A_t\ge 0\}\mathbb{I}\{r_t(\theta) \le 1+\epsilon\}
    +
    \mathbb{I}\{A_t<0\}\mathbb{I}\{r_t(\theta) \ge 1-\epsilon\},
\end{equation}
where $\mathbb{I}\{\cdot\}$ denotes the indicator function.
This advantage-sign-dependent mask is mainly used to enforce the policy-update constraint~\cite{schulman2017proximal}, preventing the policy update from becoming too large.
\subsection{Training-Inference Discrepancy, Policy Staleness, and Missing Old Logits}
\label{sec:missing_old_logits}

Under modern asynchronous LLM RL systems~\citep{fu2025areal,sheng2025laminar,sheng2025hybridflow,wang2025reinforcement}, the same parameter version can induce two distributions: the rollout distribution deployed on the inference engine, such as vLLM or SGLang, and the forward distribution deployed on the training side, such as Megatron or FSDP.
Throughout this paper, we use $\mu$ to denote the inference-side policy and $\pi$ to denote the training-side policy.
The subscript $v$ denotes the policy version, such as $\mu_v$ and $\pi_v$. By default, we use $\theta$ to denote the current version being optimized on the actor engine, and $\mathrm{old}$ to denote the rollout policy version used to generate the token on the inference engine.
Therefore, the importance ratio can be naturally decomposed into a staleness ratio $r_s$ and a discrepancy ratio between actor and rollout, $r_d$, i.e., $r_t(\theta) = r_s \times r_d$, where $r_s = \frac{\pi_\theta(y_t|x,y_{<t})}{\pi_{\mathrm{old}}(y_t|x,y_{<t})}$ and $r_d = \frac{\pi_{\mathrm{old}}(y_t|x,y_{<t})}{\mu_{\mathrm{old}}(y_t|x,y_{<t})}$.

Some works consider controlling these two terms separately, for example by masking the discrepancy ratio to mitigate the impact of numerical discrepancies~\citep{yao2025your,ma2025stabilizing}.
More recently, IcePop~\citep{zhao2025small} proposes using a strict masking threshold $c$ for the discrepancy ratio, which can be formulated as an MIS objective:
\begin{equation}\label{eq:icepop_mis}
    \mathrm{MIS}
    =
    M_{[1/c,c]}(r_d)
    \left(
        \mathbb{I}\{A_t\ge 0\}\mathbb{I}\{r_s \le 1+\epsilon\}
        +
        \mathbb{I}\{A_t<0\}\mathbb{I}\{r_s \ge 1-\epsilon\}
    \right),
\end{equation}
where we define the masking function $M_{[1/c,c]}(r):=\mathbb{I}\{1/c\le r\le c\}$.

Although this strategy has also been verified on various foundation models~\citep{xiao2026mimo,team2026kimi,zeng2026glm,wang2026ernie,team2025every}, \emph{there exists a central practical difficulty: $\pi_{\mathrm{old}}$ may be missing}.
Specifically, as shown in Figure~\ref{fig:old_logit_acquisition}, the rollout version $\mathrm{old}$ is often outdated with respect to both the behavior model and the actor model, and therefore may have already been discarded.
This is particularly common in training systems that involve partial rollout collection and asynchronous model updates~\citep{fu2025areal,sheng2025laminar,sheng2025hybridflow,wang2025reinforcement}.

In practice, existing works therefore often replace $\pi_{\mathrm{old}}$ with an approximation, for example by using a linearly interpolated policy~\cite{li20253po} or, more generally, a policy version between $\mathrm{old}$ and $\theta$ as a surrogate.
Such a decomposition does not affect the algebraic correctness of the loss function, but the two factors no longer correspond to pure discrepancy repair and pure staleness correction.
This semantic entanglement is exactly the old-logit mismatch problem.

\section{A Unified Analysis of Decoupled Correction}
\label{sec:unified_analysis}

In this section, we first provide intuition on why discrepancy repair cannot substitute for staleness correction; namely, why the decoupled approach in Eq.~\eqref{eq:icepop_mis} cannot be replaced by the standard PPO clip in Eq.~\eqref{eq:ppo_active_mask}. We then provide an analysis of existing off-policy corrections, demonstrating how they can be unified into the form of Eq.~\eqref{eq:icepop_mis}. Furthermore, we explicitly explain how the old-logit mismatch problem can lead to correction failures within the current framework. 

\subsection{Why Discrepancy Repair Cannot Substitute for Staleness Correction}
\label{sec:why_decouple}

The intuition for why the dual-side correction in Eq.~\eqref{eq:icepop_mis} cannot simply be expressed by the standard PPO correction in Eq.~\eqref{eq:ppo_active_mask} is two-fold. First, PPO primarily prevents overly large update steps by applying an asymmetric filter based on the advantage sign, whereas training-inference discrepancy repair requires a strict, symmetric constraint centered around $1$. Second, blending these decomposed terms into a single ratio forces a shared threshold, fundamentally compromising optimization. Because discrepancy repair targets numerical consistency while staleness correction controls update magnitude, they naturally demand different levels of constraint strength. A strict shared constraint stably filters out errors but severely bottlenecks learning, whereas a looser constraint accelerates early training but exposes the policy to noisy, compounded updates that increase the risk of oscillation or collapse. 

We further quantify this effect in Section~\ref{sec:threshold_tradeoff}, where exact-old-logit experiments show how discrepancy masking and PPO-CLIP still interact through the final active-token set.

\subsection{A Unified View of Existing Off-Policy Corrections}

As shown in Table~\ref{tab:ppo_variants}, existing off-policy methods in LLM RL generally decouple the optimization process into a discrepancy ratio $r_d$ and a staleness ratio $r_s$. In synchronous settings, an accessible and semantically correct old policy $\pi_{\text{old}}$ allows for an exact decomposition of training-inference discrepancy and policy staleness. However, in asynchronous RL, the latency between training and generation engines introduces an unavoidable version mismatch ($\theta \ge \text{async} \ge \text{old}$). This breaks the semantic consistency of the reference policy, corrupting the meanings of both $r_d$ and $r_s$ and causing standard decoupled corrections to fail.

\begin{table}[h]
    \centering
    \caption{Summarization of PPO variants under a unified Masked Importance Sampling (MIS) view.}
    \label{tab:ppo_variants}
    \renewcommand{\arraystretch}{1.6}
    \setlength{\tabcolsep}{4pt}
    \resizebox{\linewidth}{!}{
    \begin{tabular}{>{\small}l c c l} 
    \toprule
    \textbf{Algorithm} & \textbf{Discrepancy Ratio} $r_d$ & \textbf{Staleness Ratio} $r_s$ & \textbf{Proxy Definition} \\
    \midrule
    \multicolumn{4}{c}{
        \textbf{General Format:} 
        $ \mathrm{MIS} = M_{[1/c,c]}(r_d) \Big( \mathbb{I}_{A_t \ge 0}\mathbb{I}_{r_s \le 1+\epsilon} + \mathbb{I}_{A_t < 0}\mathbb{I}_{r_s \ge 1-\epsilon} \Big) $
    } \\
    \midrule
    PPO-clip (standard) & $1$ & $\dfrac{\pi_\theta}{\pi_{\text{old}}}$ & -- \\
    \makecell[l]{PPO-clip \& train\_infer} & $1$ & $\dfrac{\pi_\theta}{\mu_{\text{old}}}$ & -- \\
    \makecell[l]{decoupled PPO \& train\_infer} & $\dfrac{\pi_{\text{old}}}{\mu_{\text{old}}}$ & $\dfrac{\pi_\theta}{\pi_{\text{old}}}$ & -- \\
    \midrule
    PPO-EWMA & 1 & $\dfrac{\pi_\theta}{\pi_{\text{prox}}}$ & $\theta_{\text{prox}, t} = \beta \theta_{\text{prox}, t-1} + (1-\beta)\theta_t$ \\
    \makecell[l]{PPO-EWMA \& train\_infer} & $\dfrac{\pi_{\text{prox}}}{\mu_{\text{old}}}$ & $\dfrac{\pi_\theta}{\pi_{\text{prox}}}$ & $\theta_{\text{prox}, t} = \beta \theta_{\text{prox}, t-1} + (1-\beta)\theta_t$ \\
    \midrule
    linear\_prox & 1 & $\dfrac{\pi_\theta}{\pi_{\text{prox}}}$ & $\pi_{\text{prox}}=\alpha \mu_{\text{old}} + (1-\alpha)\pi_\theta$ \\
    \makecell[l]{linear\_prox \& train\_infer} & $\dfrac{\pi_{\text{prox}}}{\mu_{\text{old}}}$ & $\dfrac{\pi_\theta}{\pi_{\text{prox}}}$ & $\pi_{\text{prox}}=\alpha \mu_{\text{old}} + (1-\alpha)\pi_\theta$ \\
    \midrule
    \makecell[l]{decoupled PPO \& Async} & $\dfrac{\pi_{\text{async}}}{\mu_{\text{old}}}$ & $\dfrac{\pi_\theta}{\pi_{\text{async}}}$ & $\theta \ge \text{async} \ge \text{old}$ \\
    \bottomrule
    \end{tabular}
    }
\end{table}

To mitigate this missing reference without heavy infrastructure overhead, a common strategy is to construct a probability-space proximal policy $\pi_{\text{prox}}$ through interpolation between the current and behavior policies, such as \texttt{linear\_prox} or token-wise log-linear interpolation. However, this approach does not genuinely resolve the discrepancy, as stated below:

\vspace{1mm}
\noindent \textbf{Proposition 1.} \textit{Let $r(\theta)=\pi_\theta/\muold$. If $\pi_{\text{prox}}$ is constructed via arithmetic interpolation or token-wise log-linear interpolation, then clipping and masking on the decoupled ratios merely re-parameterizes the effective constraint boundaries of the single total ratio $r(\theta)$.
Full derivations are provided in Appendix~\ref{app:interpolation_derivation}.}
\vspace{1mm}



Because interpolation only shifts effective boundaries rather than restoring an exact reference, we explore two distinct directions to resolve the old-logit mismatch in the following sections. The first approach relies on systematic infrastructure support to directly acquire the ground-truth old logits. The second acknowledges limited system overhead and constructs a more reliable, approximate reference using a revised exponential moving average that explicitly accounts for asynchronous delays.

\section{Recovering and Approximating Old Logits}
\label{sec:recovering_old_logits}

\subsection{Exact Old-Logit Acquisition}
\label{sec:exact_old_logit_acquisition}

\begin{figure}[t]
    \centering
    \includegraphics[width=1\linewidth]{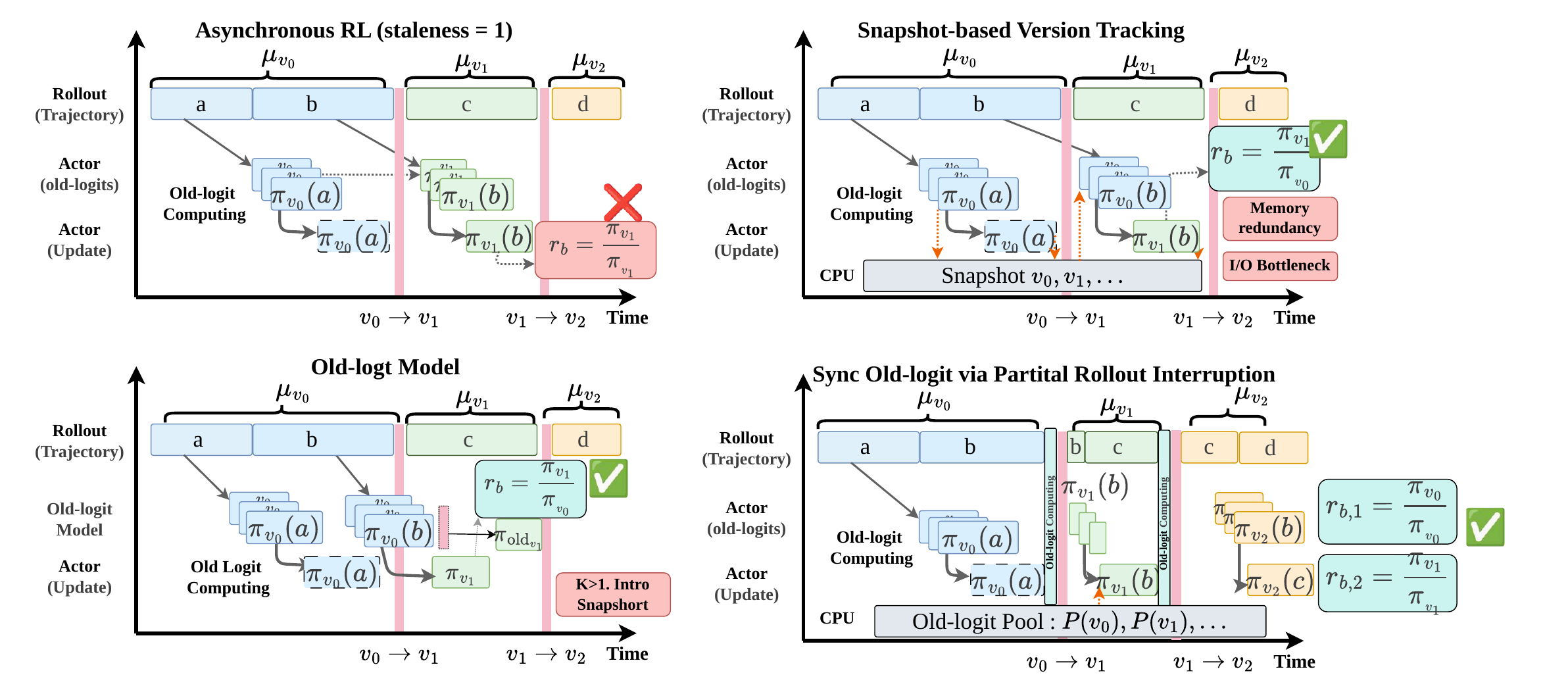}
    \caption{Exact old-logit acquisition in asynchronous RL. The top-left panel shows the original mismatch problem. The remaining panels show snapshot-based version tracking, a dedicated old-logit model, and synchronization via partial rollout interruption.}
    \label{fig:old_logit_acquisition}
\end{figure}

We first consider exact acquisition of $\pi_{\mathrm{old}}(y_t|x,y_{<t})$, the training-side token probability under the rollout version. Figure~\ref{fig:old_logit_acquisition} illustrates three possible strategies.

\paragraph{Snapshot-based version tracking.}
The most direct solution is to retain historical parameter snapshots and reload the version that generated each token or trajectory. This gives the cleanest estimate of $\pi_{\mathrm{old}}$ and therefore restores the semantic decomposition in Eq.~\eqref{eq:intro_decomposition}. Its drawback is system cost. Snapshot retention requires additional CPU or host memory, and exact recovery may require frequent actor-side version switching. With partial rollouts, a single sample can span multiple versions, which further increases switching and I/O overhead.

\paragraph{Dedicated old-logit model.}
A second option is to maintain a separate model that computes old logits while the main actor continues training. This can reduce contention on the actor path and allow overlap between old-logit computation and gradient updates. It also decouples old-logit computation from update training, so the overlap can reduce the end-to-end time of the actor stage.

\paragraph{Synchronization via partial rollout interruption.}
A third option computes old logits before a policy version disappears. Before updating parameters from version $v$ to $v+1$, the system interrupts rollout workers and returns partial trajectories. Since rollout is stopped during this interval, we can use Ray scheduling to release the rollout-side placement and temporarily switch the same resources to actor-side old-logit computation. The still-resident version $v$ is then used to compute exact old logits for the returned partial trajectories. After the old-logit pass finishes, the system switches the resources back to rollout execution and resumes generation.

This design avoids storing old weights and can provide exact logits, but it introduces synchronization stalls, resource reconfiguration overhead, and disruption to rollout parallelism. 

These three methods represent different points in a system trade-off: snapshots are exact but memory- and I/O-heavy; old-logit models enable overlap but require resource partitioning; partial interruption avoids historical storage but adds synchronization overhead.

\subsection{Revised PPO-EWMA as a Low-Cost Reference Policy}
\label{sec:ppo_ewma_method}

Exact old-logit acquisition may be too expensive for large asynchronous Agentic RL. We therefore use a (PPO-EWMA) reference policy as a low-cost approximation~\citep{hilton2022batch}. The goal is not to claim exact recovery of $\pi_{\mathrm{old}}$, but to construct a smoother reference that better tracks the center of the asynchronous version window than either the current policy or a static interpolation proxy.

PPO-EWMA maintains $\pi_{\mathrm{prox}}$ as an exponentially averaged reference policy. Given actor parameters $\theta^{(t)}$ after update step $t$, we use
\begin{equation}
    \theta_{\mathrm{prox}}^{(t)}
    =
    \frac{\sum_{k=0}^{t}\beta_{\mathrm{prox}}^{t-k}\theta^{(k)}}{\sum_{k=0}^{t}\beta_{\mathrm{prox}}^{t-k}},
    \qquad
    r_s=\frac{\pi_\theta}{\pi_{\mathrm{prox}}},
    \qquad
    r_d=\frac{\pi_{\mathrm{prox}}}{\mu_{\mathrm{old}}}.
    \label{eq:ewma_proxy_ratios}
\end{equation}
That is, we replace the unavailable $\pi_{\mathrm{old}}$ with $\pi_{\mathrm{prox}}$ for both staleness correction and discrepancy repair. This is only an approximate reference, not an exact recovery of old logits. The equivalent recursive update is provided in Appendix~\ref{app:ppo_ewma_update}.

Our adjustment is deliberately small. First, instead of using a fixed large decay, we set $\beta_{\mathrm{prox}}$ according to the expected staleness window, $\beta_{\mathrm{prox}}\approx W_{\mathrm{stale}}/(W_{\mathrm{stale}}+2)$. This places the EWMA reference near the middle of the asynchronous version window and prevents it from lagging behind the rollout queue. Appendix~\ref{app:ppo_ewma_update} gives the center-of-mass derivation.

Second, we add an automatic reset to avoid accumulating excessively stale versions in the EWMA reference. As $\theta_{\mathrm{prox}}$ averages over more historical actor states, it may drift away from the policy version used by recent rollouts. This makes the discrepancy ratio $\pi_{\mathrm{prox}}/\mu_{\mathrm{old}}$ deviate further from one and causes the Train-Infer Mask to reject many tokens. We therefore monitor the Train-Infer Mask value $\rho_t$, defined as the fraction of tokens that remain active after discrepancy masking and PPO clipping. When $\rho_t<\tau$, where $\tau$ is set to $0.9$ in our experiments unless otherwise stated, we reset
\begin{equation}
    \theta_{\mathrm{prox}}^{(t)} \leftarrow \theta^{(t)} .
\end{equation}
This clears stale history and re-centers the proxy reference around the current actor. Section~\ref{sec:ewma_ablation} provides a more detailed analysis of this behavior.

\section{Experiments}
\label{sec:experiments}

\subsection{Experimental Setup}
\label{sec:experimental_setup}

We evaluate Agentic RL tasks using two representative policy backbones: the dense \texttt{Qwen3-4B} model and the MoE \texttt{Qwen3-30B-A3B} model. Training data are drawn from an \href{https://huggingface.co/datasets/guanzhong2/TU_Pipeline}{Agentic RL corpus} spanning multiple environments. Evaluation covers the retail, airline, and telecom domains of $\tau^2$-Bench~\citep{barres2025tau}, together with the in-store and delivery splits of VitaBench~\citep{he2025vitabench}. For both benchmarks, we report task-level average success and pass metrics.

To isolate the effect studied in this paper, we adopt an asynchronous RL setup with explicit control over the maximum version gap between rollout workers and the actor, which we cap at three. We eliminate additional staleness from minibatch reuse and multiple PPO epochs where possible, ensuring that the observed staleness is primarily due to rollout--training asynchrony. Detailed hyperparameters, resource configurations, and environment descriptions are provided in Appendix~\ref{app:experimental_details}.

We consider several practical methods: \textbf{Decoupled PPO}, which employs the decoupled objective but relies on the available asynchronous reference rather than the true $\pi_{\text{old}}$ policy; \textbf{Linear\_prox}, a lightweight linear interpolation-based proximal policy; and \textbf{PPO-EWMA}, our enhanced EWMA reference with staleness-aware decay and optional reset. Additionally, we include \textbf{Snapshot}, which recovers exact old logits via true version tracking. This is feasible in our setup because we control the maximum version gap.

\subsection{Main Results}
\label{sec:main_results}

Table~\ref{tab:main_results} compares exact old-logit recovery, proxy references, and PPO-EWMA on held-out Agentic benchmarks. Snapshot$^\dagger$ serves as an idealized reference because it assumes exact old logits are available. PPO-EWMA is the practical method we emphasize: it consistently improves over Decoupled PPO and Linear\_prox, and often approaches the Snapshot results.

\begin{table*}[t]
\caption{Main results on the held-out Agentic benchmark suite for the dense \texttt{Qwen3-4B} model and the \texttt{Qwen3-30B-A3B} MoE model. VitaBench is separated into in-store and delivery splits. Snapshot$^\dagger$ denotes an idealized setting where exact old logits are available. The best performance among practical methods (first three rows) is \textbf{bolded}.}
\centering
\scriptsize
\setlength{\tabcolsep}{3pt}
\resizebox{\textwidth}{!}{
\begin{tabular}{ll|cc|cc|cc|cc|cc}
\toprule
\textbf{Backbone} & \textbf{Method}
& \multicolumn{2}{c|}{\textbf{$\tau^2$-Bench (Retail)}}
& \multicolumn{2}{c|}{\textbf{$\tau^2$-Bench (Airline)}}
& \multicolumn{2}{c|}{\textbf{$\tau^2$-Bench (Telecom)}}
& \multicolumn{2}{c|}{\textbf{VitaBench (In-store)}}
& \multicolumn{2}{c}{\textbf{VitaBench (Delivery)}} \\
\cmidrule(lr){3-4} \cmidrule(lr){5-6} \cmidrule(lr){7-8} \cmidrule(lr){9-10} \cmidrule(lr){11-12}
& & \textbf{avg@4} & \textbf{pass@4}
& \textbf{avg@4} & \textbf{pass@4}
& \textbf{avg@2} & \textbf{pass@2}
& \textbf{avg@2} & \textbf{pass@2}
& \textbf{avg@2} & \textbf{pass@2} \\
\midrule
\texttt{Qwen3-4B} & Decoupled PPO & 63.96 & 88.60 & 53.5 & 72 & 40 & 50 & 19.83 & 37 & 19.56 & 33 \\
\texttt{Qwen3-4B} & Linear\_prox   & 64.40 & 86.84 & \textbf{54} & 72 & 37.5 & 50 & 22.37 & 40 & 19.10 & 28 \\
\texttt{Qwen3-4B} & PPO-EWMA       & \textbf{65.72} & \textbf{90.35} & \textbf{54} & \textbf{74} & \textbf{42.5} & \textbf{52.5} & \textbf{25} & \textbf{50} & \textbf{25.88} & \textbf{39} \\
\midrule
\texttt{Qwen3-4B} & Snapshot$^\dagger$ & {66.23} & 89.47 & {56} & {76} & {42.5} & {52.5} & {28.89} & 47 & {27.33} & {42} \\
\midrule
\makecell[l]{\texttt{Qwen3-30B-A3B}} & Decoupled PPO & 65.43 & 89.47 & 57 & 76 & 44.75 & 55 & 18.28 & 32 & 25.88 & 39 \\
\makecell[l]{\texttt{Qwen3-30B-A3B}} & Linear\_prox   & 65.8 & 87.7 & 53.5 & 74 & 44 & 55 & 31.47 & 47  & 20.74 & 33 \\
\makecell[l]{\texttt{Qwen3-30B-A3B}} & PPO-EWMA       & \textbf{67.82} & \textbf{92.1} & \textbf{60} & \textbf{82} & \textbf{45} & \textbf{57.5} & \textbf{33.41} & \textbf{48} & \textbf{28.49} & \textbf{43} \\
\midrule
\makecell[l]{\texttt{Qwen3-30B-A3B}} & Snapshot$^\dagger$ & {69.70} & {92.1} & 59 & 80 & {45} & {57.5} & {34.62} & {50} & {30.74} & {45} \\

\bottomrule
\end{tabular}
}
\label{tab:main_results}
\end{table*}

On the dense 4B model, PPO-EWMA obtains the best pass@4 on retail and the best pass@2 on VitaBench in-store, while tying the best telecom scores. On the 30B MoE model, it is strongest on the airline split and ties the best retail pass@4 and telecom avg@2. These results suggest that a maintained EWMA reference can recover much of the benefit of exact correction without requiring exact old-logit recovery.

\subsection{System Overhead of Exact Old-Logit Acquisition}
\label{sec:system_overhead_exact_old_logits}

Table~\ref{tab:exact_old_logit_runtime_overhead} summarizes the system cost of exact old-logit acquisition. Snapshot is accurate but expensive: it requires version switching, snapshot storage, and extra recovery time, with the cost becoming much larger on the 30B MoE model.

PPO-EWMA is much cheaper because it only maintains a lightweight proxy reference. In the left subtable, its CPU storage and extra time are far lower than Snapshot for both backbones. The right subtable shows that a dedicated old-logit model can overlap part of the computation, but its benefit depends on the resource partition ratio and still requires additional model-side infrastructure.


\begin{table*}[t]
\caption{System-overhead measurement protocols for exact old-logit acquisition. Snapshot recovery mainly adds version-switch latency and CPU storage; a dedicated old-logit model changes the overlapped time allocation.}
\centering
\makebox[\textwidth][c]{%
\begin{subtable}[t]{0.36\textwidth}
\caption{Snapshot vs PPO-EWMA.}
\vspace{0pt}
\centering
\scriptsize
\setlength{\tabcolsep}{5pt}
\begin{tabular}{lcc|cc}
\toprule
\textbf{Metric} & \textbf{4B} & \textbf{30B} & \textbf{4B} & \textbf{30B} \\
\midrule
Switch latency (s) & 7 & 14.2  & 7 & 14.2 \\
CPU storage (GB) & 40 & 76.4 & 7.9 & 15.2 \\
Extra time (s) & 25 & 150 & 8 & 34 \\
\bottomrule
\end{tabular}
\label{tab:snapshot_overhead}
\end{subtable}
\hspace{0.07\textwidth}
\begin{subtable}[t]{0.48\textwidth}
\caption{Dedicated old-logit model.}
\vspace{0pt}
\centering
\scriptsize
\setlength{\tabcolsep}{3pt}
\begin{tabular}{lccccc}
\toprule
\textbf{Ratio} & \textbf{Old-logit (s)} & \textbf{Update (s)} & \textbf{Single (s)} & \textbf{Overlap (s)} & \textbf{Change (\%)} \\
\midrule
1:2 & 243 & 272 & 305 & 284 & -6.8 \\
1:3 & 243 & 206 & 237 & 254 & +7.17 \\
\bottomrule
\end{tabular}
\label{tab:old_logit_model_overhead}
\end{subtable}
}

\label{tab:exact_old_logit_runtime_overhead}
\end{table*}

Overall, Table~\ref{tab:exact_old_logit_runtime_overhead} supports the same conclusion as Table~\ref{tab:main_results}: exact recovery is useful as a strong reference, but PPO-EWMA offers a better performance--cost trade-off for practical asynchronous training.

\subsection{Threshold Trade-off under Exact Old Logits}
\label{sec:threshold_tradeoff}

Using exact old logits from Snapshot, we study how the discrepancy and stale-policy thresholds affect optimization. All threshold-tradeoff curves in this section are collected with the \texttt{Qwen3-4B} backbone. In each run name, the first number denotes the discrepancy threshold and the second denotes the stale-policy threshold; for example, \texttt{snap1003\_1006} corresponds to $1.003$ and $1.006$.

\begin{figure}[t]
    \centering
    \begin{subfigure}[t]{0.55\linewidth}
        \centering
        \includegraphics[width=\linewidth]{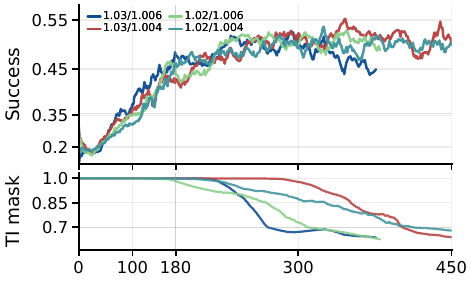}
        \caption{Threshold trade-off.}
        \label{fig:threshold_tradeoff_main}
    \end{subfigure}
    \hfill
    \begin{subfigure}[t]{0.43\linewidth}
        \centering
        \includegraphics[width=\linewidth]{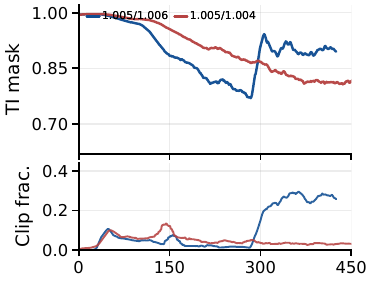}
        \caption{Mask--clip interaction.}
        \label{fig:mask_clip_interaction}
    \end{subfigure}
    \caption{Exact-old-logit threshold analysis. Each label reports the discrepancy and stale-policy thresholds.}
    \label{fig:threshold_and_mask_clip}
\end{figure}

Figure~\ref{fig:threshold_and_mask_clip} summarizes the trade-off. A looser discrepancy threshold keeps more tokens active and improves early learning, but it also admits more biased off-policy tokens and can destabilize later trajectories. A stricter threshold slows the beginning but yields smoother retained-token dynamics and reduces late collapse. Appendix~\ref{sec:threshold_appendix_examples} shows the same pattern for nearby settings.

Panel~\ref{fig:mask_clip_interaction} further shows that discrepancy masking and PPO clipping are coupled. With the same discrepancy threshold, a looser stale-policy threshold lets more problematic tokens survive early, causing the Train-Infer Mask to drop lower; later, stronger PPO clipping caps the remaining updates and helps the mask recover.

\subsection{PPO-EWMA Analysis and Ablation}
\label{sec:ewma_ablation}

We next study the revised PPO-EWMA through three training-time signals: Task success, Train-Infer Mask, and PPO-CLIP ratio. The core issue is reference-policy lag: a useful EWMA reference must be smooth enough to stabilize updates, but not so stale that it collapses the Train-Infer Mask. We therefore use a staleness-aware decay and add automatic reset to re-center the reference when the mask becomes too low. Detailed decay and threshold ablations are provided in Appendix~\ref{app:ewma_additional_ablations}.

\begin{figure}[t]
    \centering
    \includegraphics[width=0.96\linewidth]{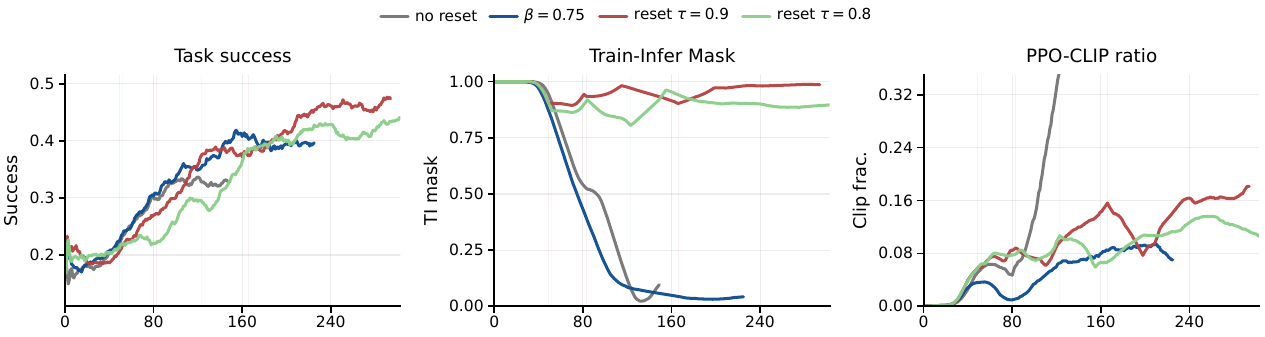}
    \caption{Effect of automatic reset for $\beta=0.75$. Resetting the EWMA reference when the Train-Infer Mask becomes too low prevents late-stage collapse. Vertical lines mark reset events.}
    \label{fig:ewma_autoreset}
\end{figure}

Figure~\ref{fig:ewma_autoreset} shows the main result. Without reset, the EWMA reference can drift far enough that Train-Infer Mask collapses. With reset, only a few re-centering events are needed to recover a high Train-Infer Mask value while preserving most of the early Task success gain of $\beta=0.75$. This supports the practical design: use EWMA as a low-cost approximate reference, but reset it when the proxy becomes too stale.

\section{Conclusion}
\label{sec:conclusion}

We identified and studied the missing-old-logit problem in asynchronous Agentic RL for LLMs, which undermines decoupled off-policy correction. We address it from two directions: exact old-logit recovery at the infrastructure level, and low-cost approximation with a revised PPO-EWMA reference. Our results show that exact recovery improves correction fidelity, while PPO-EWMA provides a practical alternative when exact recovery is too expensive. A limitation is that PPO-EWMA remains an approximate reference rather than a true reconstruction of the missing old policy, and may become unreliable under highly non-stationary staleness or extreme version gaps.

\bibliographystyle{plain}
\bibliography{reference}

\appendix

\section{Detailed Derivations for Interpolation-Based Proxies}
\label{app:interpolation_derivation}

This appendix gives the derivation behind Proposition~1. We write the total ratio between the current training policy and the behavior-side reference as:
\begin{equation}
    r_\theta = \frac{\pi_\theta(y|x)}{\muold(y|x)} .
\end{equation}
For a proximal policy $\pi_{\mathrm{prox}}$, the decoupled objective applies a discrepancy mask to 
\begin{equation}
    r_d = \frac{\pi_{\mathrm{prox}}(y|x)}{\muold(y|x)}
\end{equation}
and PPO clipping to 
\begin{equation}
    r_s = \frac{\pi_\theta(y|x)}{\pi_{\mathrm{prox}}(y|x)} .
\end{equation}
For simplification, we consider the threshold for masking as $\epsilon_1$ and $\epsilon_2$, instead of $c$, i.e., the masked importance sampling is:
\begin{equation}
    \mathrm{MIS} = M_{[1-\epsilon_1, 1+\epsilon_1]}(r_d) \left( \mathbb{I}\{A_t \ge 0\}\mathbb{I}\{r_s \le 1+\epsilon_2\} + \mathbb{I}\{A_t < 0\}\mathbb{I}\{r_s \ge 1-\epsilon_2\} \right) .
\end{equation}
The key question is whether constructing $\pi_{\mathrm{prox}}$ by interpolation introduces a genuinely new reference, or whether it only changes the effective constraints on $r_\theta$. We use $\alpha$ for the behavior-policy weight, consistent with Table~\ref{tab:ppo_variants}.

\paragraph{Log-linear interpolation.}
Consider the token-wise log-linear interpolation:
\begin{equation}
    \log \widetilde{\pi}_{\mathrm{prox}}(y|x) = \alpha \log \muold(y|x) + (1-\alpha)\log \pi_\theta(y|x), \quad \alpha \in (0,1).
\end{equation}
In this case, the discrepancy ratio becomes:
\begin{equation}
    r_d = \frac{\widetilde{\pi}_{\mathrm{prox}}}{\muold} = \left(\frac{\pi_\theta}{\muold}\right)^{1-\alpha} = r_\theta^{1-\alpha}.
\end{equation}
The mask condition $1-\epsilon_1 \le r_d \le 1+\epsilon_1$ restricts $r_\theta$ to the interval:
\begin{equation}
    (1-\epsilon_1)^{1/(1-\alpha)} \le r_\theta \le (1+\epsilon_1)^{1/(1-\alpha)} .
\end{equation}
The PPO-side ratio is $r_s = \pi_\theta / \widetilde{\pi}_{\mathrm{prox}} = r_\theta^\alpha$. We analyze the constraints based on the sign of $A_t$:
\begin{itemize}
    \item \textbf{Case $A_t \ge 0$}: The clip requires $r_s \le 1+\epsilon_2$, implying $r_\theta \le (1+\epsilon_2)^{1/\alpha}$. Combined with the mask, the active region for $r_\theta$ is:
    \begin{equation}
        (1-\epsilon_1)^{1/(1-\alpha)} \le r_\theta \le \min\left( (1+\epsilon_1)^{1/(1-\alpha)}, (1+\epsilon_2)^{1/\alpha} \right) .
    \end{equation}
    \item \textbf{Case $A_t < 0$}: The clip requires $r_s \ge 1-\epsilon_2$, implying $r_\theta \ge (1-\epsilon_2)^{1/\alpha}$. Combined with the mask, the active region is:
    \begin{equation}
        \max\left( (1-\epsilon_1)^{1/(1-\alpha)}, (1-\epsilon_2)^{1/\alpha} \right) \le r_\theta \le (1+\epsilon_1)^{1/(1-\alpha)} .
    \end{equation}
\end{itemize}
First-order expansion shows $\epsilon_{1,\mathrm{eff}} \approx \frac{\epsilon_1}{1-\alpha}$ and $\epsilon_{2,\mathrm{eff}} \approx \frac{\epsilon_2}{\alpha}$. This shows that log-linear interpolation merely re-parameterizes the active region of the original ratio $r_\theta$.

\paragraph{Arithmetic interpolation.}
Consider the linear proxy:
\begin{equation}
    \pi_{\mathrm{prox}}(y|x) = \alpha\muold(y|x) + (1-\alpha)\pi_\theta(y|x), \quad \alpha \in (0,1).
\end{equation}
The discrepancy ratio is $r_d = \alpha + (1-\alpha) r_\theta$, so the mask condition $1-\epsilon_1 \le r_d \le 1+\epsilon_1$ becomes:
\begin{equation}
    1 - \frac{\epsilon_1}{1-\alpha} \le r_\theta \le 1 + \frac{\epsilon_1}{1-\alpha} .
\end{equation}
The PPO-side ratio is $r_s = \frac{r_\theta}{\alpha + (1-\alpha)r_\theta}$. Again, we analyze by the sign of $A_t$:
\begin{itemize}
    \item \textbf{Case $A_t \ge 0$}: Solving $r_s \le 1+\epsilon_2$ for $r_\theta$ gives:
    \begin{equation}
        r_\theta \le \frac{\alpha(1+\epsilon_2)}{1-(1-\alpha)(1+\epsilon_2)} .
    \end{equation}
    The active region is the intersection of the lower mask bound and the tighter of the two upper bounds.
    \item \textbf{Case $A_t < 0$}: Solving $r_s \ge 1-\epsilon_2$ for $r_\theta$ gives:
    \begin{equation}
        r_\theta \ge \frac{\alpha(1-\epsilon_2)}{1-(1-\alpha)(1-\epsilon_2)} .
    \end{equation}
    The active region is the intersection of the upper mask bound and the tighter of the two lower bounds.
\end{itemize}
Applying $(1-x)^{-1} \approx 1+x$ reveals an effective radius $\epsilon_{2,\mathrm{eff}} \approx \frac{\epsilon_2}{\alpha}$. Thus, arithmetic interpolation also functions as a set of scaled constraints on the single total ratio $r_\theta$.

\paragraph{Numerical effective thresholds.}
Table~\ref{tab:interpolation_effective_thresholds} instantiates the above mappings for version gaps $n\in\{1,2,3\}$, where the behavior-policy weight is $\alpha=1/(n+1)$. We include the thresholds used in our experiments and several nearby settings. The clip intervals are allowed to be asymmetric because our implementation uses separate lower and upper PPO-CLIP thresholds; the symmetric PPO interval is a special case. Across these settings, interpolation changes the effective mask and clip ranges on the same total ratio rather than introducing an independent old-policy reference.

\begin{table}[t]
\centering
\small
\setlength{\tabcolsep}{4.5pt}
\renewcommand{\arraystretch}{1.18}
\begin{tabular}{cccccc}
\toprule
\textbf{Original mask} & \textbf{Original clip} & \textbf{$n$} & \textbf{Interpolation} & \textbf{Mask on $r_d$} & \textbf{Clip on $r_s$} \\
\midrule
$[0.990,1.010]$ & $[0.997,1.004]$ & 1 & Linear & $[0.9800,1.0200]$ & $[0.9940,1.0080]$ \\
$[0.990,1.010]$ & $[0.997,1.004]$ & 1 & Log-linear & $[0.9801,1.0201]$ & $[0.9940,1.0080]$ \\
$[0.990,1.010]$ & $[0.997,1.004]$ & 2 & Linear & $[0.9850,1.0150]$ & $[0.9911,1.0121]$ \\
$[0.990,1.010]$ & $[0.997,1.004]$ & 2 & Log-linear & $[0.9850,1.0150]$ & $[0.9910,1.0120]$ \\
$[0.990,1.010]$ & $[0.997,1.004]$ & 3 & Linear & $[0.9867,1.0133]$ & $[0.9881,1.0162]$ \\
$[0.990,1.010]$ & $[0.997,1.004]$ & 3 & Log-linear & $[0.9867,1.0134]$ & $[0.9881,1.0161]$ \\
\midrule
$[0.995,1.005]$ & $[0.997,1.004]$ & 1 & Linear & $[0.9900,1.0100]$ & $[0.9940,1.0080]$ \\
$[0.995,1.005]$ & $[0.997,1.004]$ & 1 & Log-linear & $[0.9900,1.0100]$ & $[0.9940,1.0080]$ \\
$[0.995,1.005]$ & $[0.997,1.004]$ & 2 & Linear & $[0.9925,1.0075]$ & $[0.9911,1.0121]$ \\
$[0.995,1.005]$ & $[0.997,1.004]$ & 2 & Log-linear & $[0.9925,1.0075]$ & $[0.9910,1.0120]$ \\
$[0.995,1.005]$ & $[0.997,1.004]$ & 3 & Linear & $[0.9933,1.0067]$ & $[0.9881,1.0162]$ \\
$[0.995,1.005]$ & $[0.997,1.004]$ & 3 & Log-linear & $[0.9933,1.0067]$ & $[0.9881,1.0161]$ \\
\midrule
$[0.990,1.010]$ & $[0.996,1.006]$ & 1 & Linear & $[0.9800,1.0200]$ & $[0.9920,1.0121]$ \\
$[0.990,1.010]$ & $[0.996,1.006]$ & 1 & Log-linear & $[0.9801,1.0201]$ & $[0.9920,1.0120]$ \\
$[0.990,1.010]$ & $[0.996,1.006]$ & 2 & Linear & $[0.9850,1.0150]$ & $[0.9881,1.0182]$ \\
$[0.990,1.010]$ & $[0.996,1.006]$ & 2 & Log-linear & $[0.9850,1.0150]$ & $[0.9880,1.0181]$ \\
$[0.990,1.010]$ & $[0.996,1.006]$ & 3 & Linear & $[0.9867,1.0133]$ & $[0.9842,1.0244]$ \\
$[0.990,1.010]$ & $[0.996,1.006]$ & 3 & Log-linear & $[0.9867,1.0134]$ & $[0.9841,1.0242]$ \\
\midrule
$[0.980,1.020]$ & $[0.997,1.004]$ & 1 & Linear & $[0.9600,1.0400]$ & $[0.9940,1.0080]$ \\
$[0.980,1.020]$ & $[0.997,1.004]$ & 1 & Log-linear & $[0.9604,1.0404]$ & $[0.9940,1.0080]$ \\
$[0.980,1.020]$ & $[0.997,1.004]$ & 2 & Linear & $[0.9700,1.0300]$ & $[0.9911,1.0121]$ \\
$[0.980,1.020]$ & $[0.997,1.004]$ & 2 & Log-linear & $[0.9702,1.0301]$ & $[0.9910,1.0120]$ \\
$[0.980,1.020]$ & $[0.997,1.004]$ & 3 & Linear & $[0.9733,1.0267]$ & $[0.9881,1.0162]$ \\
$[0.980,1.020]$ & $[0.997,1.004]$ & 3 & Log-linear & $[0.9734,1.0268]$ & $[0.9881,1.0161]$ \\
\bottomrule
\end{tabular}
\renewcommand{\arraystretch}{1.0}
\caption{Effective thresholds on the total ratio $r_\theta=\pi_\theta/\muold$ induced by interpolation-based proxies. The first block corresponds to the thresholds used in our experiment. For version gap $n$, we set the behavior-policy weight $\alpha=1/(n+1)$.}
\label{tab:interpolation_effective_thresholds}
\end{table}

\begin{figure}[t]
    \centering
    \includegraphics[width=0.76\linewidth]{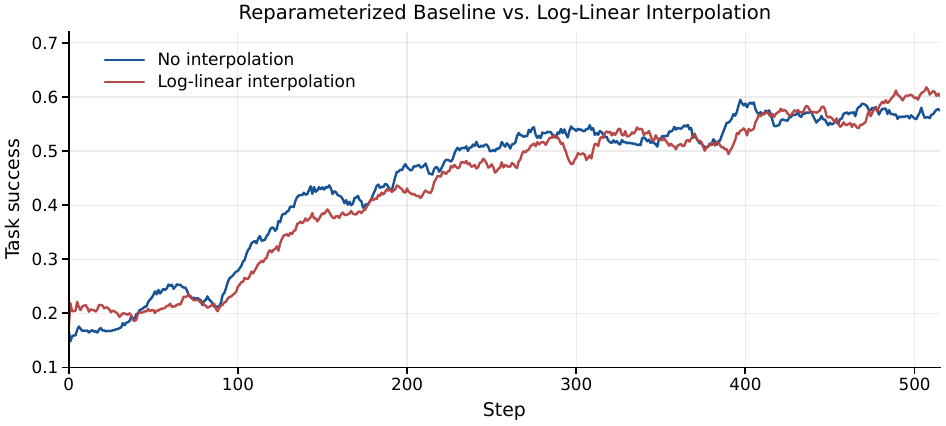}
    \caption{Training curves for the reparameterized no-interpolation baseline and the log-linear interpolation proxy. The two curves are nearly indistinguishable over training, consistent with the derivation that log-linear interpolation mainly changes the ratio parameterization.}
    \label{fig:reference_reparameterization_comparison}
\end{figure}

Figure~\ref{fig:reference_reparameterization_comparison} provides an empirical check of the same reparameterization view. After removing the interpolation and applying the corresponding reparameterized constraints directly, the training curve is almost identical to the log-linear interpolation run. This supports the interpretation that the interpolation proxy mainly changes the effective ratio boundary rather than introducing an independent correction effect.

These derivations explain why interpolation-based proxies can act as useful stabilizers but should not be interpreted as exact training--inference discrepancy repair. The ratios remain deterministic transformations of $r_\theta=\pi_\theta/\muold$; no term reconstructs the missing training-side old policy logits.

\section{PPO-EWMA Update Details}
\label{app:ppo_ewma_update}

This appendix records the standard PPO-EWMA parameterization used in Section~\ref{sec:ppo_ewma_method}. It is included for completeness; the main text focuses on the two asynchronous adjustments, namely staleness-aware decay selection and auto-reset.

Let $\theta^{(t)}$ be the actor parameters after update step $t$, and let $\beta_{\mathrm{prox}}\in(0,1)$ be the EWMA decay. We maintain a cumulative normalization weight $w^{(t)}$ and proximal parameters $\theta_{\mathrm{prox}}^{(t)}$ as
\begin{equation}
    w^{(t)} = 1 + \beta_{\mathrm{prox}} w^{(t-1)},
    \qquad
    \theta_{\mathrm{prox}}^{(t)}
    =
    \frac{1}{w^{(t)}}\theta^{(t)}
    +
    \beta_{\mathrm{prox}}\frac{w^{(t-1)}}{w^{(t)}}\theta_{\mathrm{prox}}^{(t-1)} .
    \label{eq:ewma_recursive_appendix}
\end{equation}
Unrolling the recursion gives the closed-form expression used in the main text:
\begin{equation}
    \theta_{\mathrm{prox}}^{(t)}
    =
    \frac{\sum_{k=0}^{t}\beta_{\mathrm{prox}}^{t-k}\theta^{(k)}}{\sum_{k=0}^{t}\beta_{\mathrm{prox}}^{t-k}} .
    \label{eq:ewma_closed_form_appendix}
\end{equation}
Thus, $\theta_{\mathrm{prox}}^{(t)}$ is a normalized exponentially weighted average of historical actor states. A larger $\beta_{\mathrm{prox}}$ keeps a longer memory of previous policies, while a smaller value places more weight on the current actor.

The decay factor also determines the effective look-back depth of the reference. In the stationary limit, the center of mass of the EWMA history is
\begin{equation}
    \mathrm{COM}_{\mathrm{prox}}
    =
    \frac{\sum_{k=0}^{\infty} k\beta_{\mathrm{prox}}^k}{\sum_{k=0}^{\infty}\beta_{\mathrm{prox}}^k}
    =
    \frac{\beta_{\mathrm{prox}}}{1-\beta_{\mathrm{prox}}}.
    \label{eq:ewma_com_appendix}
\end{equation}
For an asynchronous version window of width $W_{\mathrm{stale}}$, aligning this center of mass with the midpoint of the window gives
\begin{equation}
    \frac{\beta_{\mathrm{prox}}}{1-\beta_{\mathrm{prox}}}
    \approx
    \frac{W_{\mathrm{stale}}}{2},
    \qquad
    \beta_{\mathrm{prox}} \approx \frac{W_{\mathrm{stale}}}{W_{\mathrm{stale}}+2}.
    \label{eq:ewma_beta_rule_appendix}
\end{equation}

\section{Full PPO Variant Comparison}
\label{app:full_variant_table}

Table~\ref{tab:full_ppo_variants} provides the expanded comparison behind
Table~\ref{tab:ppo_variants}. We distinguish two ratios whenever possible:
$r_1$ is the policy-update or staleness ratio used by PPO-style clipping, while
$r_2$ is the training--inference discrepancy or proxy-reference ratio. For
positive advantages, PPO clipping keeps samples before the upper clipping
boundary; for negative advantages, it keeps samples after the lower clipping
boundary. The discrepancy constraint, when used, is symmetric around one and
does not depend on the advantage sign.

\begin{table*}[t]
\centering
\caption{Expanded comparison of PPO variants and proxy-policy corrections. The
columns $A_+$ and $A_-$ show the active regions for positive and negative
advantages, respectively.}
\label{tab:full_ppo_variants}
\scriptsize
\setlength{\tabcolsep}{3pt}
\renewcommand{\arraystretch}{1.25}
\resizebox{\textwidth}{!}{
\begin{tabular}{>{\small}l c >{\small}c >{\small}c >{\small}c}
\toprule
\textbf{Algorithm} & \textbf{Ratio Formula} & \textbf{$A_+$} & \textbf{$A_-$} & \textbf{Role} \\
\midrule
PPO-clip (standard)
& $r_1=\dfrac{\pi_\theta(a_t|s_t)}{\pi_{\mathrm{old}}(a_t|s_t)}$
& $r_1\in(0,1+\epsilon_1)$
& $r_1\in(1-\epsilon_1,\infty)$
& $M(r_1)$ \\
\midrule
\makecell[l]{PPO-clip \\ \& train\_infer}
& $r_1=\dfrac{\pi_\theta(a_t|s_t)}{\mu_{\mathrm{old}}(a_t|s_t)}$
& $r_1\in(0,1+\epsilon_1)$
& $r_1\in(1-\epsilon_1,\infty)$
& $M(r_1)$ \\
\midrule
\makecell[l]{decoupled PPO \\ \& train\_infer}
& \makecell{$r_1=\dfrac{\pi_\theta(a_t|s_t)}{\pi_{\mathrm{old}}(a_t|s_t)}$ \\[4pt]
            $r_2=\dfrac{\pi_{\mathrm{old}}(a_t|s_t)}{\mu_{\mathrm{old}}(a_t|s_t)}$}
& \makecell{$r_1\in(0,1+\epsilon_1)$ \\ $r_2\in(1\pm\epsilon_2)$}
& \makecell{$r_1\in(1-\epsilon_1,\infty)$ \\ $r_2\in(1\pm\epsilon_2)$}
& $M(r_2)\cdot M(r_1)$ \\
\midrule
PPO-EWMA
& \makecell{$r_1=\dfrac{\pi_\theta(a_t|s_t)}{\pi_{\mathrm{prox}}(a_t|s_t)}$ \\[4pt]
            $r_2=\dfrac{\pi_{\mathrm{prox}}(a_t|s_t)}{\mu_{\mathrm{old}}(a_t|s_t)}$ \\[4pt]
            $\theta_{\mathrm{prox},t}=\beta\theta_{\mathrm{prox},t-1}+(1-\beta)\theta_t$}
& $r_1\in(0,1+\epsilon_1)$
& $r_1\in(1-\epsilon_1,\infty)$
& $r_2\cdot M(r_1)$ \\
\midrule
\makecell[l]{PPO-EWMA \\ \& train\_infer}
& \makecell{$r_1=\dfrac{\pi_\theta(a_t|s_t)}{\pi_{\mathrm{prox}}(a_t|s_t)}$ \\[4pt]
            $r_2=\dfrac{\pi_{\mathrm{prox}}(a_t|s_t)}{\mu_{\mathrm{old}}(a_t|s_t)}$ \\[4pt]
            $\theta_{\mathrm{prox},t}=\beta\theta_{\mathrm{prox},t-1}+(1-\beta)\theta_t$}
& \makecell{$r_1\in(0,1+\epsilon_1)$ \\ $r_2\in(1\pm\epsilon_2)$}
& \makecell{$r_1\in(1-\epsilon_1,\infty)$ \\ $r_2\in(1\pm\epsilon_2)$}
& $M(r_2)\cdot r_2\cdot M(r_1)$ \\
\midrule
linear\_prox
& \makecell{$r_1=\dfrac{\pi_\theta(a_t|s_t)}{\pi_{\mathrm{prox}}(a_t|s_t)}$ \\[4pt]
            $r_2=\dfrac{\pi_{\mathrm{prox}}(a_t|s_t)}{\mu_{\mathrm{old}}(a_t|s_t)}$ \\[4pt]
            $\pi_{\mathrm{prox}}=\alpha\mu_{\mathrm{old}}+(1-\alpha)\pi_\theta$}
& \makecell{$r_1\in(0,1+\epsilon_1)$ \\ $r_2\in(1\pm\epsilon_2)$}
& \makecell{$r_1\in(1-\epsilon_1,\infty)$ \\ $r_2\in(1\pm\epsilon_2)$}
& $r_2\cdot M(r_1)$ \\
\midrule
\makecell[l]{linear\_prox \\ \& train\_infer}
& \makecell{$r_1=\dfrac{\pi_\theta(a_t|s_t)}{\pi_{\mathrm{prox}}(a_t|s_t)}$ \\[4pt]
            $r_2=\dfrac{\pi_{\mathrm{prox}}(a_t|s_t)}{\mu_{\mathrm{old}}(a_t|s_t)}$ \\[4pt]
            $\pi_{\mathrm{prox}}=\alpha\mu_{\mathrm{old}}+(1-\alpha)\pi_\theta$}
& $r_1\in(0,1+\epsilon_1)$
& $r_1\in(1-\epsilon_1,\infty)$
& $M(r_2)\cdot r_2\cdot M(r_1)$ \\
\midrule
\makecell[l]{decoupled PPO \\ \& train\_infer \\ \& Async}
& \makecell{$r_1=\dfrac{\pi_\theta(a_t|s_t)}{\pi_{\mathrm{async}}(a_t|s_t)}$ \\[4pt]
            $r_2=\dfrac{\pi_{\mathrm{async}}(a_t|s_t)}{\mu_{\mathrm{old}}(a_t|s_t)}$ \\[4pt]
            $\theta\ge\mathrm{async}\ge\mathrm{old}$}
& \makecell{$r_1\in(0,1+\epsilon_1)$ \\ $r_2\in(1\pm\epsilon_2)$}
& \makecell{$r_1\in(1-\epsilon_1,\infty)$ \\ $r_2\in(1\pm\epsilon_2)$}
& $M(r_2)\cdot M(r_1)$ \\
\bottomrule
\end{tabular}
}
\renewcommand{\arraystretch}{1.0}
\end{table*}

The standard PPO-clip objective uses a single ratio and therefore cannot
separate policy staleness from training--inference discrepancy. The
train\_infer variant replaces the denominator with the rollout-side behavior
distribution, but still applies PPO clipping to a mixed ratio. Decoupled PPO
restores the intended two-factor structure when the training-side old policy
$\pi_{\mathrm{old}}$ is available. In asynchronous training, however, this
reference may be missing; using $\pi_{\mathrm{async}}$ or a constructed
$\pi_{\mathrm{prox}}$ then gives a practical proxy rather than an exact
decomposition. This is why proxy methods can stabilize optimization while still
leaving the semantic old-logit mismatch unresolved.

\section{Additional Threshold Examples}
\label{sec:threshold_appendix_examples}

We provide additional threshold comparisons to show that the trade-off in Section~\ref{sec:threshold_tradeoff} is not specific to one pair of hyperparameters. As in the main text, the first number in each run name is the training--inference discrepancy threshold and the second number is the stale-policy threshold. The upper panel of each success figure reports the task success curve, and the lower panel reports the difference between the two curves. These differences make the early-speed and late-stability trade-off easier to inspect.

\begin{figure}[t]
    \centering
    \includegraphics[width=0.72\linewidth]{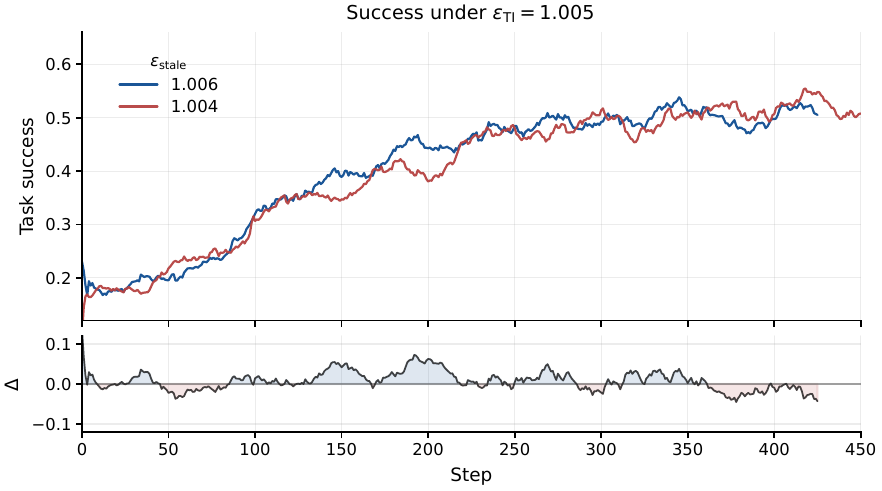}
    \caption{Additional threshold comparison: \texttt{snap1005\_1006} vs.\ \texttt{snap1005\_1004}. The two runs share the same discrepancy threshold and differ only in stale-policy control. The looser stale-policy threshold gives a faster early trajectory, while the stricter setting catches up more smoothly later.}
    \label{fig:threshold_extra_1005}
\end{figure}

Figure~\ref{fig:threshold_extra_1005} isolates the effect of changing the stale-policy threshold under a fixed discrepancy threshold of $1.005$. The looser threshold initially retains more update signal and therefore improves early progress. Later, however, the gap narrows and the curve becomes more oscillatory, consistent with the interpretation that early under-filtering creates additional correction pressure in later training.

\begin{figure}[t]
    \centering
    \includegraphics[width=0.72\linewidth]{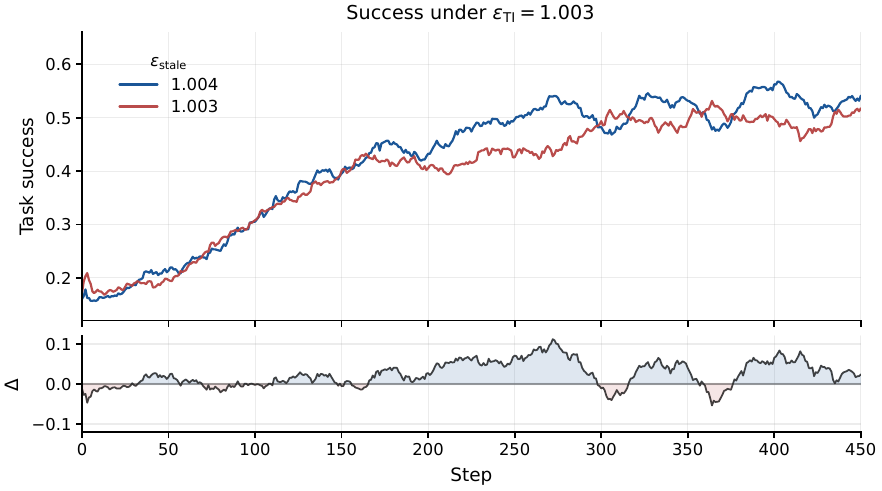}
    \caption{Additional threshold comparison: \texttt{snap1003\_1004} vs.\ \texttt{snap1003\_1003}. The same early-speed and late-stability trade-off appears under a stricter discrepancy threshold.}
    \label{fig:threshold_extra_1003}
\end{figure}

Figure~\ref{fig:threshold_extra_1003} repeats the comparison with discrepancy threshold $1.003$. Since this discrepancy filter is already stricter, both settings discard more biased tokens than the $1.005$ pair. The remaining difference is therefore smaller, but the same qualitative pattern remains: the looser stale-policy setting improves early learning speed, while the stricter setting gives a more controlled trajectory.

\begin{figure}[t]
    \centering
    \includegraphics[width=0.72\linewidth]{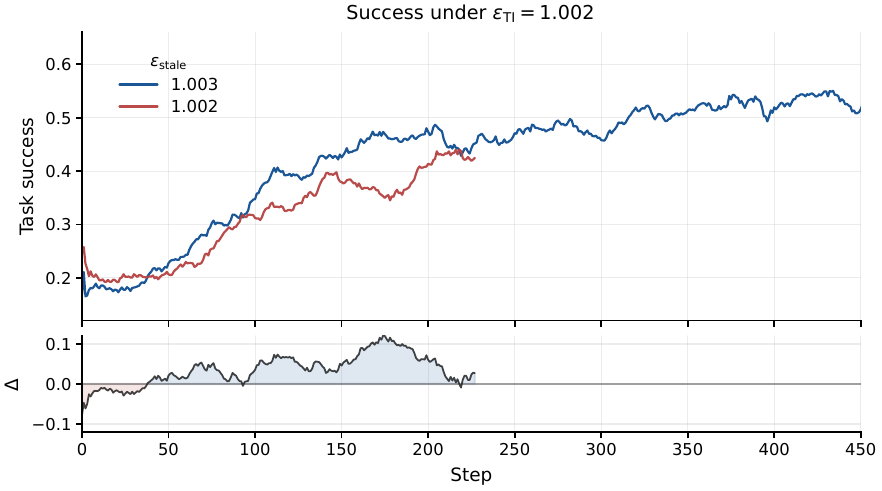}
    \caption{Additional threshold comparison: \texttt{snap1002\_1003} vs.\ \texttt{snap1002\_1002}. Under the strictest discrepancy threshold, the retained signal is smaller and learning is slower, but the training trajectory is more stable.}
    \label{fig:threshold_extra_1002}
\end{figure}

Figure~\ref{fig:threshold_extra_1002} uses discrepancy threshold $1.002$, which filters the most aggressively among the appendix examples. This setting illustrates the other side of the trade-off: stronger filtering reduces the number of biased tokens and stabilizes training, but it also limits the amount of useful off-policy signal available for policy improvement. The result is slower optimization and a lower early success curve.

\begin{figure}[t]
    \centering
    \includegraphics[width=0.72\linewidth]{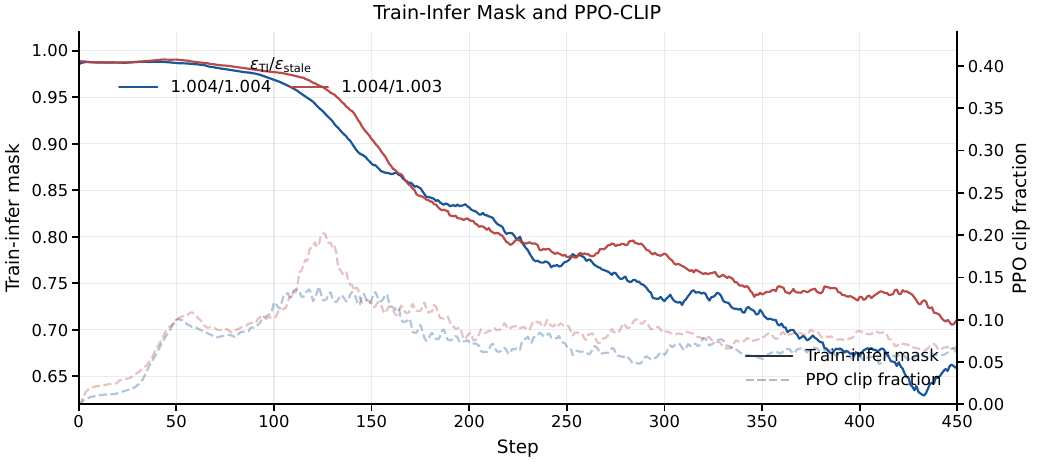}
    \caption{Additional interaction example: \texttt{snap1004\_1004} vs.\ \texttt{snap1004\_1003}. The discrepancy mask and PPO-CLIP activation change together even though the exact old logits are available.}
    \label{fig:mask_clip_interaction_extra}
\end{figure}

Figure~\ref{fig:mask_clip_interaction_extra} provides a second example of the coupling between discrepancy repair and PPO-CLIP. The two runs share discrepancy threshold $1.004$ and differ in the stale-policy threshold. Changing the stale-policy threshold changes not only the PPO clip fraction, but also the trajectory of the training--inference mask. This confirms that the two mechanisms interact through the active-token set: a looser stale-policy constraint can leave more questionable tokens for discrepancy repair, while stronger discrepancy filtering changes how often PPO-CLIP becomes active on the remaining tokens. Thus, exact old-logit recovery restores the correct semantic decomposition, but threshold tuning still has to account for the joint behavior of the two constraints.

\section{Additional PPO-EWMA Ablations}
\label{app:ewma_additional_ablations}

This appendix provides the detailed PPO-EWMA ablations behind Section~\ref{sec:ewma_ablation}. We use three diagnostic quantities: Task success, Train-Infer Mask, and PPO-CLIP ratio. Task success measures optimization progress, while Train-Infer Mask and PPO-CLIP ratio show whether the EWMA reference remains usable as a proximal policy.

\paragraph{Decay-factor behavior.}
\begin{figure}[t]
    \centering
    \includegraphics[width=0.96\linewidth]{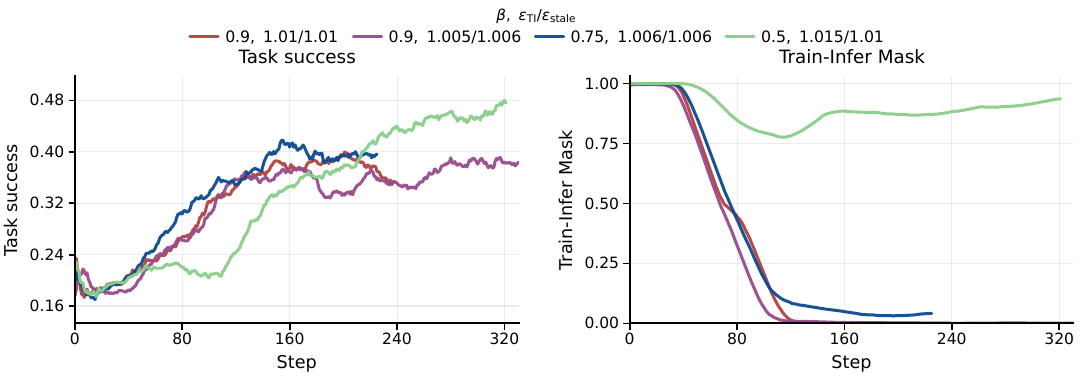}
    \caption{PPO-EWMA decay comparison. A large decay can accumulate stale reference history and collapse the Train-Infer Mask, while the staleness-aware decay gives faster early Task success.}
    \label{fig:ewma_beta_comparison}
\end{figure}

Figure~\ref{fig:ewma_beta_comparison} compares $\beta=0.9$, $\beta=0.75$, and $\beta=0.5$. When $\beta=0.9$, the EWMA reference has a long memory and remains strongly affected by early actor versions. As training proceeds, $\pi_{\mathrm{prox}}$ becomes increasingly misaligned with recent rollout policies, so the discrepancy ratio $\pi_{\mathrm{prox}}/\mu_{\mathrm{old}}$ drifts away from one. This makes the Train-Infer Mask progressively more aggressive. Tightening either the discrepancy threshold or the PPO-CLIP ratio threshold does not remove this failure mode: both $\beta=0.9$ runs eventually reach an almost zero Train-Infer Mask value. This supports our interpretation that the issue is not only threshold choice, but also reference-policy lag.

The theoretically motivated $\beta=0.75$ setting aligns the EWMA center of mass with the middle of the asynchronous version window. It therefore gives the fastest early Task success increase among the compared decays. However, without any reset, this run also suffers a late Train-Infer Mask collapse: the minimum Train-Infer Mask value falls below $2\%$, and the final Train-Infer Mask value remains only about $3\%$. In contrast, $\beta=0.5$ has a shorter memory and keeps many more tokens active, but its Task success curve is less efficient early. These results show the core trade-off of PPO-EWMA: a longer memory provides a stronger stabilizing anchor, but can accumulate stale-policy bias; a shorter memory adapts faster, but weakens the proximal reference.

\begin{figure}[t]
    \centering
    \includegraphics[width=0.96\linewidth]{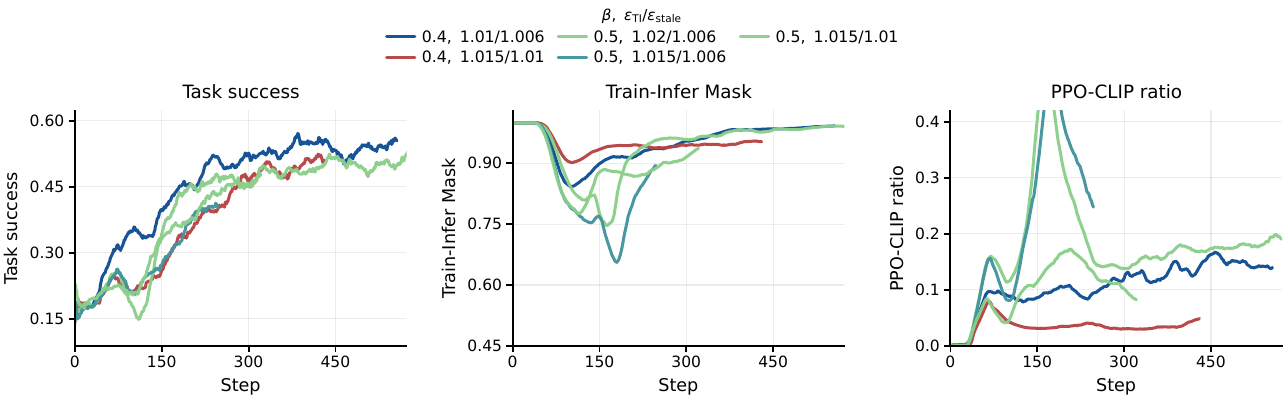}
    \caption{PPO-EWMA threshold interaction for $\beta=0.4$ and $\beta=0.5$. Looser Train-Infer Mask or PPO-CLIP ratio thresholds can improve early progress, but they also admit more noisy off-policy updates and may cause a mid-training success drop. The coupled mask and clip dynamics can later recover part of the trajectory by filtering or capping the problematic updates.}
    \label{fig:ewma_threshold_interaction}
\end{figure}

Figure~\ref{fig:ewma_threshold_interaction} shows that threshold choice remains important even when the EWMA decay is small enough to avoid the severe collapse observed for $\beta=0.9$. For $\beta=0.4$ and $\beta=0.5$, looser Train-Infer Mask or PPO-CLIP ratio thresholds retain more update signal early and can produce faster initial gains. However, these looser settings also allow more mismatched tokens to participate in optimization, which can cause a visible mid-training success drop. The subsequent recovery is consistent with the dual-constraint interpretation: discrepancy masking removes tokens whose Train-Infer ratio becomes too unreliable, while PPO-CLIP caps the update magnitude on the remaining tokens.

\paragraph{Automatic reset behavior.}
Figure~\ref{fig:ewma_autoreset} shows that automatic reset resolves the late-collapse failure while preserving most of the early efficiency of $\beta=0.75$. The reset mechanism is conservative: in the two runs with thresholds $\tau=0.9$ and $\tau=0.8$, it is triggered only three and two times, respectively. Nevertheless, these few resets are sufficient to clear the stale history accumulated in $\theta_{\mathrm{prox}}$ and recover a high Train-Infer Mask value. This indicates that the most important effect of reset is early re-centering of the proxy reference; after the reference is corrected, later training usually remains in a healthier region without frequent interventions.

The reset indicator further confirms that reset is not acting as a continuously active stabilizer. Most later steps have zero consecutive low-ratio streaks and no reset trigger. Therefore, the improvement should be interpreted as a small number of re-centering events that remove stale EWMA history after the reference has drifted too far, rather than as frequent intervention throughout training.

\section{Experimental Details}
\label{app:experimental_details}

\paragraph{Evaluation splits.}
For $\tau^2$-Bench, we use all available \texttt{base} samples for the retail and airline domains. For the telecom domain, we use the \texttt{test} split.

\paragraph{Implementation framework.}
We use ROLL as the base training framework for our experiments. ROLL provides a
well-structured agent abstraction. 


\section{Additional System Measurements}
\label{app:additional_system_measurements}

\subsection{Snapshot Old-Logit Overhead}
\label{app:snapshot_overhead_details}

Table~\ref{tab:appendix_snapshot_overhead_details} reports the detailed overhead introduced by enabling snapshot-based old-logit computation. Both runs use \texttt{old\_log\_probs\_source=snapshot\_old} with asynchronous snapshot recovery enabled. The 4B run contains 77 measured training steps before termination, while the 30B-A3B run contains 11 measured steps. These numbers should therefore be interpreted as system-cost measurements from the observed run windows rather than final training results.

\begin{table}[t]
\centering
\small
\setlength{\tabcolsep}{4.5pt}
\renewcommand{\arraystretch}{1.12}
\begin{tabular}{lcc}
\toprule
\textbf{Metric} & \textbf{Qwen3-4B} & \textbf{Qwen3-30B-A3B} \\
\midrule
Measured steps & 77 & 11 \\
Snapshot size (GB) & 8.04 & 15.29 \\
Max resident snapshots & 5 & 5 \\
Max resident snapshot storage (GB) & 40.22 & 76.43 \\
Snapshot save time / step (s) & 3.95 & 6.07 \\
Snapshot load time (s) & 7.11 & 14.22 \\
Snapshot restore time (s) & 3.01 & 5.66 \\
Historical forward time (s) & 45.05 & 147.18 \\
Total snapshot-old stage time / step (s) & 95.20 & 178.22 \\
Version shards / batch & 1.81 & 1.09 \\
Historical version shards / batch & 1.74 & 1.00 \\
Max staleness gap & 3 & 3 \\
\bottomrule
\end{tabular}
\renewcommand{\arraystretch}{1.0}
\caption{Detailed overhead from snapshot-based old-logit computation. The dominant cost is the additional historical-model forward pass used to reconstruct old logits. Snapshot retention also adds substantial resident storage pressure, especially for the 30B-A3B model.}
\label{tab:appendix_snapshot_overhead_details}
\end{table}

The main additional cost is not snapshot saving itself, but the combination of loading or restoring historical versions and running extra forward passes to compute old logits. In the 4B run, snapshot-old computation adds roughly 95 seconds per measured step on average. In the 30B-A3B run, the same stage adds roughly 178 seconds per measured step. The 30B-A3B setting also keeps up to 76.43GB of resident snapshot state when five snapshots are retained.

Both runs eventually terminate with Ray actor unavailability after the Raylet exits unexpectedly. The logs do not prove that snapshot recovery is the sole cause, but the measured resident snapshot storage and additional forward time show that exact old-logit recovery materially increases memory pressure and actor-side compute load.

\section{Limitations}
\label{app:limitations}

This work has several limitations. First, our experiments are conducted on
representative dense and MoE backbones, but we have not yet validated the
proposed correction methods on models at the several-hundred-billion-parameter
scale. At that scale, memory pressure, communication overhead, expert routing
behavior, and rollout--training scheduling can become qualitatively different
from the settings studied here. As a result, the empirical trade-offs observed
in our current experiments may not fully capture the behavior of extremely large
industrial training runs.

Second, our infrastructure analysis focuses on the main sources of overhead for
exact old-logit acquisition, including snapshot retention, version switching,
historical forward computation, and partial rollout interruption. However, we do
not provide a fine-grained end-to-end systems study of all infrastructure
components. In particular, we have not exhaustively analyzed scheduler behavior,
network communication, placement-group fragmentation, memory swapping, rollout
worker idleness, or failure recovery under sustained large-scale asynchronous
training.

Third, the throughput measurements in this paper should be interpreted as
indicative rather than comprehensive. They are sufficient to show that exact
old-logit recovery introduces non-trivial system cost, but we have not performed
a large-scale throughput sweep across many cluster sizes, model scales, sequence
lengths, rollout lengths, and staleness windows. A more complete systems
evaluation would be needed to precisely characterize the throughput frontier and
to identify the best engineering design for each deployment regime.


\end{document}